\documentclass[acmsmall,screen]{acmart}
\usepackage{subfigure}
\renewcommand\footnotetextcopyrightpermission[1]{} 
\AtBeginDocument{%
  \providecommand\BibTeX{{%
    \normalfont B\kern-0.5em{\scshape i\kern-0.25em b}\kern-0.8em\TeX}}}

\setcopyright{acmcopyright}
\copyrightyear{2018}
\acmYear{2018}
\acmDOI{10.1145/1122445.1122456}

\acmJournal{JACM}
\acmVolume{37}
\acmNumber{4}
\acmArticle{111}
\acmMonth{8}

\usepackage{indentfirst}
\usepackage{color}
\usepackage{url}

\begin{document}

\title{Attribute Controllable Beautiful Caucasian Face Generation by Aesthetics Driven Reinforcement Learning}

\author{Xin Jin}
\affiliation{%
	\institution{Beijing Electronic Science and Technology Institute}
	\city{Beijing}
	\state{Beijing}
	\country{China}
	\postcode{100070}
}
\author{Shu Zhao}
\affiliation{%
	\institution{Beijing Electronic Science and Technology Institute}
	\city{Beijing}
	\state{Beijing}
	\country{China}
	\postcode{100070}
}
\author{Le Zhang}
\authornote{Corresponding author email: lezhang.thu@gmail.com}
\affiliation{%
	\institution{Beijing Electronic Science and Technology Institute}
	\city{Beijing}
	\state{Beijing}
	\country{China}
	\postcode{100070}
}
\author{Xin Zhao}
\affiliation{%
	\institution{Beijing Electronic Science and Technology Institute}
	\city{Beijing}
	\state{Beijing}
	\country{China}
	\postcode{100070}
}
\author{Qiang Deng}
\affiliation{%
	\institution{Beijing Electronic Science and Technology Institute}
	\city{Beijing}
	\state{Beijing}
	\country{China}
	\postcode{100070}
}
\author{Chaoen Xiao}
\affiliation{%
	\institution{Beijing Electronic Science and Technology Institute}
	\city{Beijing}
	\state{Beijing}
	\country{China}
	\postcode{100070}
}



\begin{abstract}
	In recent years, image generation has made great strides in improving the quality of images, producing high-fidelity ones. Also, quite recently, there are architecture designs, which enable GAN to unsupervisedly learn the semantic attributes represented in different layers. However, there is still a lack of research on generating face images more consistent with human aesthetics. Based on EigenGAN [He et al., ICCV 2021], we build the techniques of reinforcement learning into the generator of EigenGAN. The agent tries to figure out how to alter the semantic attributes of the generated human faces towards more preferable ones. To accomplish this, we trained an aesthetics scoring model that can conduct facial beauty prediction. We also can utilize this scoring model to analyze the correlation between face attributes and aesthetics scores. Empirically, using off-the-shelf techniques from reinforcement learning would not work well. So instead, we present a new variant incorporating the ingredients emerging in the reinforcement learning communities in recent years. Compared to the original generated images, the adjusted ones show clear distinctions concerning various attributes. Experimental results using the MindSpore, show the effectiveness of the proposed method. Altered facial images are commonly more attractive, with significantly improved aesthetic levels.
\end{abstract}


\begin{teaserfigure}
	\includegraphics[width=\textwidth]{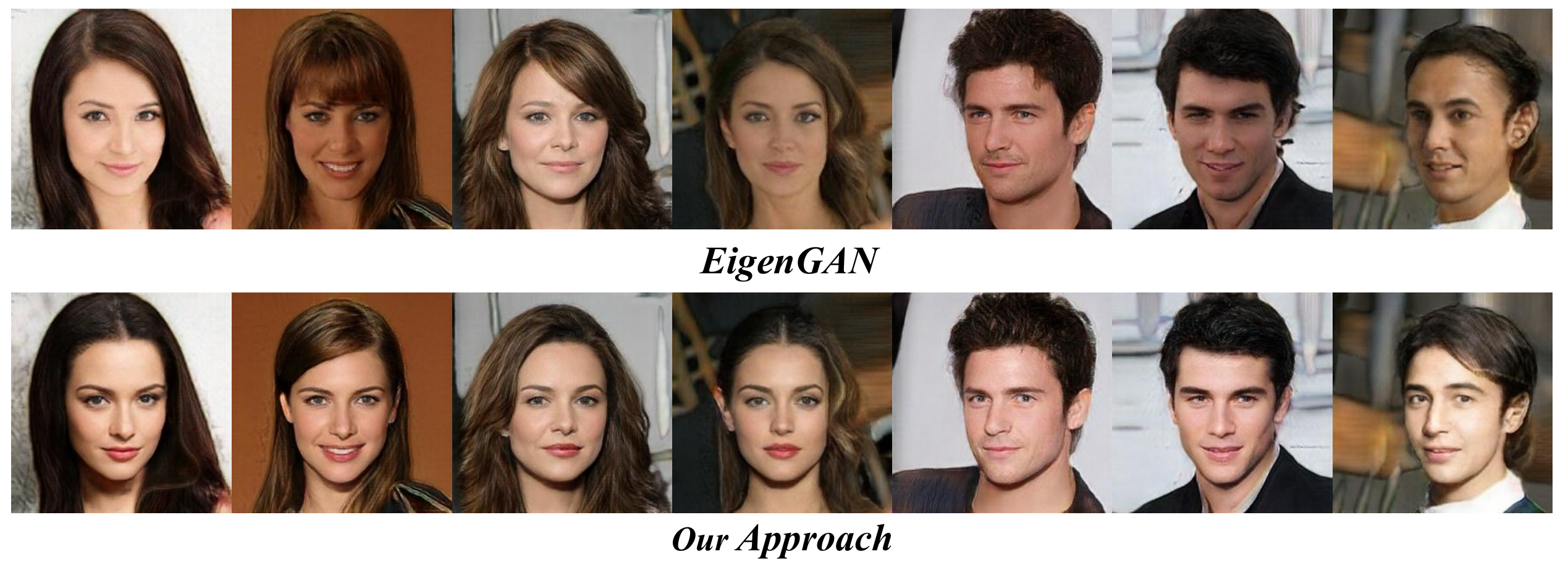}
	\caption{Top: the original faces generated by EigenGAN \cite{2021EigenGAN}. Bottom: aesthetically improved ones by aesthetics-driven reinforcement learning over semantic attributes of the generator.}
	\label{fig:teaser}
\end{teaserfigure}
\maketitle

\section{Introduction}
In recent years, Generative Adversarial Networks (GANs) \cite{goodfellow2014generative} show promising performance on image synthesis. A parallel line of research tries to learn interpretable GAN generators, which can discover explicit dimensions controlling the semantic attributes. InfoGAN in \cite{chen2016infogan} learns disentangled representations by decomposing the conventional single unstructured noise vector into incompressible noise and a set of structured latent variables. Motivated by style transfer literature, Karras et al. proposed StyleGAN in \cite{karras2019style}, enabling separations of high-level attributes like pose and identity. Most recently, He et al. proposed EigenGAN  \cite{2021EigenGAN}, which for each generator layer directly embeds in one linear subspace model with an orthogonal basis. During the training, EigenGAN automatically learns a set of ``eigen-dimensions'' corresponding to interpretable semantic attributes. So until now, people have several ways to manipulate the semantics within the GANs, or said in other words, semantically edit them. One problem followed is how to enhance the generated face images positively. 

In this paper, we incorporate the techniques of reinforcement learning (RL) into the attribute values searching in EigenGAN. The first requirement of the RL agent is the design of the reward function. The objective is to enhance the images to the ones more aesthetically attractive. So for this, we trained an aesthetics score network based on SCUT-FBP5500 released in \cite{liang2018scut}. By this, we give a multi-attribute model using MindSpore\cite{mindspore}, which can conduct automatic Caucasian face beautification for the generated images, as shown in Fig.~\ref{fig:teaser}. The user can select different facial attributes in the interaction section to change the generated image.

\section{Related Work}
\label{sec:relate}

\textbf{Reinforcement Learning in Aesthetics.} Li et al. proposed an RL-based framework named Fast A3RL (Aesthetics-Aware Adversarial) \cite{2018A2RL,2019A3RL}, which dedicates to the task of image cropping. 
This task requires the agent to extract a well-composed region from the input image. Li et al. first formulated the cropping as a decision-making process. Next, they introduced an RL-based cropping network, which contains an actor-critic architecture \cite{mnih2016asynchronous}. Although Fast A3RL is also an application of RL in aesthetics, the work proposed here differs in several aspects. On one side, Fast A3RL relies on a specially designed reward function, which involves human-designed heuristics besides an aesthetics score network. The solution discussed here, however, only requires a straightforward regression over SCUT-FBP5500 \cite{liang2018scut} to give facial beauty predictions. The solution does not need any hand-craft features. On the other side, RL techniques utilized in Fast A3RL are conventional policy gradient methods. So it requires two networks, i.e., a policy network and a value network. Whereas we follow the approach taken in the literature on image captioning \cite{rennie2017self,pan2020x,zhang2021vinvl}, and use a self-critical baseline.  This form not only simplifies the network design but also liberates the agent from learning accurate value estimations.

\textbf{ Facial Attractiveness Enhancement.} There are other facial image beautification techniques proposed in recent years. Diamant et al. proposed 
Beholder-GAN \cite{2019Beholder-GAN}, which is a variant of PGGAN (Progressive Growing) \cite{karras2017progressive} and CGAN (Conditional) \cite{mirza2014conditional}. Diamant et al. first trained a GAN based on beauty scores. Then for real images, the latent vectors and the beauty scores are recovered using GAN inversion techniques. To make input images more attractive, one only needs to increase the beauty scores recovered. Although effective, Beholder-GAN struggles to preserve the facial identity of the input image. In \cite{2020BeautifyBasedOnGAN}, Zhou and Xiao proposed an approach based on InterFaceGAN \cite{2020InterFaceGAN}, and treated the ``beauty'' as a binary semantic. We note that in InterFaceGAN, the main assumption is ``for any binary semantic
(eg, male v.s. ~female), there exists a hyperplane in the
latent space serving as the separation boundary.'' Although intuitively implausible (as beauty is not obvious binary semantic as other semantics like the presence of eyeglasses or not), Zhou and Xiao managed to validate the existence of such a beauty hyper-plane. Like InterFaceGAN, they can now semantically edit this beauty semantic of input images, generating aesthetically more attractive ones. Besides, by inserting an identity preserving loss, they proposed an enhanced version of Beholder-GAN. Nonetheless, as presented (and illustrated) in their paper, this enhanced version only solved identity-preserving weakness to some extend. We note that for these approaches, GANs are still acting like a black box. Whereas, by contrast, we seek for precise controlling of explicit semantics. The pre-trained model in \cite{2020InterFaceGAN} is not released, so the comparison to it is not conducted in the experiment.

\begin{figure*}
	\centering
	\includegraphics[width=0.98\textwidth]{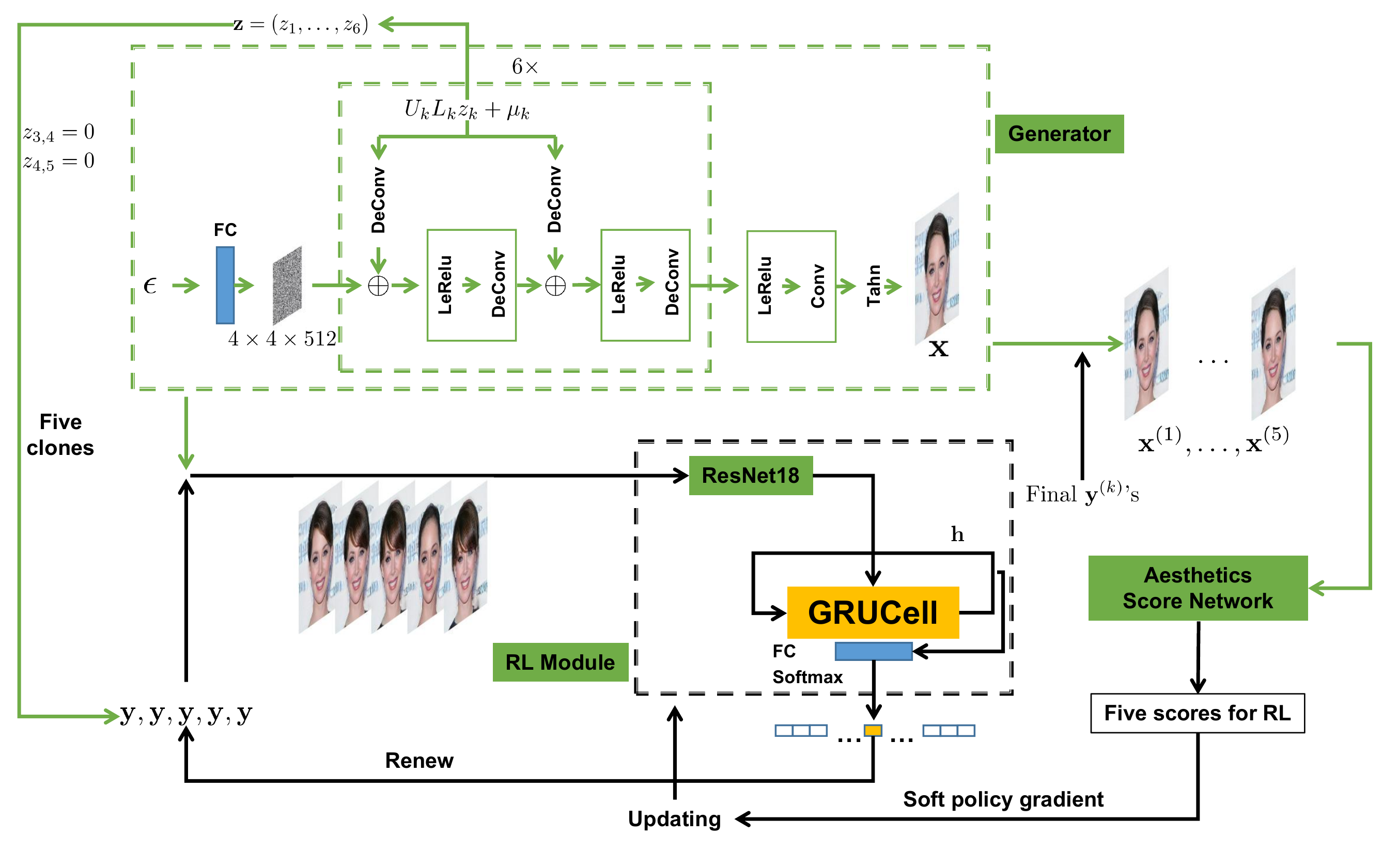}
	\caption{Illustration of the training procedure over two semantics $z_{3,4}$ and $z_{4,5}$. After setting $z_{3,4}$ and $z_{4,5}$ to zero, by cloning $\mathbf{z}$, we have five identical $\mathbf{y}$'s. We generate five images by feeding $\mathbf{y}$'s into the generator. We use ResNet18 to extract the features of the five images, and input them into a GRUCell. Next, the hidden state of GRUCell is input into a fully connected layer, and a softmax layer then gives the action distribution for choosing values for $z_{3,4}$. We renew these newly chosen values into $\mathbf{y}$'s. Again, by feeding these updated $\mathbf{y}$'s into the generator, we have five \textbf{new fresh} images. The procedure repeats, and we get the sampling values for $z_{4,5}$. After updating the values for $z_{4,5}$ into $\mathbf{y}$'s, we get the final vectors $\mathbf{y}^{(k)}$'s.  
		We again feed $\mathbf{y}^{(k)}$'s into the generator to get the final five images corresponding to five randomly sampled trajectories.
		The aesthetics score network will score them and return five scores. These scores provide an informative baseline for the soft policy gradient. The RL module is next updated to raise the probabilities for the trajectories with higher scores. Finally, for the inference, the agent will not sample actions after the softmax layer. In contrast, it greedily chooses the most probable one.}
	\label{fig:nn_structure}
\end{figure*}

\section{EigenGAN + RL}
In this section, we detail the building blocks for incorporating the policy gradient algorithm into the generator of EigenGAN. We first briefly review the preliminaries of RL and the architecture of the generator of EigenGAN. We depict it as part of Fig.~\ref{fig:nn_structure}.

\subsection{Preliminaries for RL}
In RL, the agent constantly interacts with the environment, and this process is described as a Markov Decision Process (MDP). We define the MDP as $(\mathcal{S},\mathcal{A},P,r,\rho_0,\gamma)$, where $\mathcal{S}$ and $\mathcal{A}$ are the state and action space resp., $P:\mathcal{S}\times \mathcal{A}\times \mathcal{S}\to \mathbb{R}$ is the state transition probability, $r:\mathcal{S}\to\mathbb{R}$ is the reward function, $\rho_0$ is the distribution of the initial state $s_0$, and $\gamma$ is the discounting factor satisfying $0\le\gamma\le 1$.

The objective is to learn a stochastic policy $\pi:\mathcal{S}\times\mathcal{A}\to [0,1]$, such that when taking actions $\mathbf{a}_t\sim\pi(\mathbf{a}_t|\mathbf{s}_t)$, the expected reward of the trajectories is maximized, i.e.,
\begin{equation}
	\pi^{*}=\underset{\pi}{\arg \max } \mathbb{E}_{\mathbf{s}_0,\mathbf{a}_0,\dotsc}\left[\gamma^tr(\mathbf{s}_t)\right]\text{,}
\end{equation}
where $\mathbf{s}_0\sim\rho_0(\mathbf{s}_0)$, $\mathbf{a}_t\sim\pi(\mathbf{a}_t|\mathbf{s}_t)$, $\mathbf{s}_{t+1}\sim P(\mathbf{s}_{t+1}|\mathbf{s}_t,\mathbf{a}_t)$. 

In this paper, the circumstance considered is a simplified version of the one above. The discounting factor $\gamma$ equals 1. Also, no rewards exist for intermediate states. The agent only gets a reward for the terminal state. The reward value is the aesthetic score of the image after a series of adjustments.

\subsection{Generator of EigenGAN}
The controlling parameters of the generator consist of not only the conventionally noise input $\mathbf{\epsilon}\in \mathcal{N}(0,\mathbf{I})$, but also a sequence of latent variables $\mathbf{z}_i\in\mathbb{R}^q$, which are similarly sampled from $\mathcal{N}_q(0,\mathbf{I})$ (for EigenGAN, $q$ is set to 6). Six blocks of transposed convolution layers constitute the main part of the neural architecture. Strides of the two transposed convolutions are 2 and 1 resp. Hence, the size of the image is scaled by a factor of 2 when traveling through one such block. Within the block, the randomly sampled point (due to the randomness of $\mathbf{z}_i$) from the subspace model $S_i=(\mathbf{U}_i,\mathbf{L}_i,\mathbf{\mu}_i)$ is embedded into the image by element-wise addition. For the notations, $\mathbf{U}_i$ is the orthonormal basis, $\mathbf{L}_i$ is a diagonal matrix indicating the ``importance,'' and $\mathbf{\mu}_i$ is the origin of the subspace.

For the input part, it is a fully connected layer mapping $\mathbf{\epsilon}$ into a $4\times 4\times 512$ image. Correspondingly for the output, it is passed into a convolutional layer, getting a generated $256\times 256$ image of 3 channels.

\subsection{Learn to set semantic attributes}
\label{sec:rl_method}
For EigenGAN, dimensions from latent variables $\mathbf{z}_i$ correspond to interpretable or controllable semantic attributes. As an example, the fourth dimension of $\mathbf{z}_3$ is related to the control of the bangs. The sixth dimension of $\mathbf{z}_5$, for a specific pre-trained generator, corresponds to the lipstick color. So as we move along these dimensions, the generator produces a sequence of images with continuous changes corresponding to the relevant semantic attributes. Six layers in EigenGAN brings us six such $\mathbf{z}$'s. Recall every $\mathbf{z}_i\in\mathbb{R}^q$, and $q=6$. We only adopt a subset of these controllable semantic attributes, as the others are a bit entangled or drastic, like the first dimension of $\mathbf{z}_3$ is possible to be unsupervisedly trained to change the age or even the gender of the generated face.
The premise of the adjustments is to keep the unaltered version still recognizable.
Let $C$ be the set of selected controllable semantic attributes. We only discuss how to let the RL agent learn to choose the preferred values. We defer the details of the RL algorithm to the next section.

Formally as shown in \ref{fig:nn_structure} (for conciseness, only two controllable dimensions are involved), we first set the dimensions corresponding to the attributes in $C$ as zero. Any other ``sensible'' values (EigenGAN samples these values according to a standard normal distribution) can also be okay, as the agent will set or override these values anyway.  The agent learns how to select over this dimension during the training process. After these setting zero operations, we concatenate these altered $\mathbf{z}_i$'s into one single vector. Let it be $\mathbf{y}$. We clone the single vector five times. The value of five caters to the five trajectories or decision sequences sampled by the RL agent. Intuitively, we train the agent over these sampled trajectories. The agent then gradually learns to select the preferred one.

As depicted in Fig.~\ref{fig:nn_structure}, these five (identical) $\mathbf{y}$'s will be passed into the generator. Accompanied with the same $\mathbf{\epsilon}$, these five $\mathbf{y}$'s induce five (identical) images. Remind that for a frozen pre-trained EigenGAN generator, passing in the same $\mathbf{\epsilon}$ and $\mathbf{z}_i$'s deterministic-ally generates the same image. The next step is the conventional feature extraction process. In this paper, we directly use a pre-trained ResNet18 model, for which we replace the last fully connected layer with a counterpart of output dimension 512. The value of 512 is consistent with the length of $\mathbf{\epsilon}$. We also choose this value as the default dimension of latent vectors. We do not freeze the other layers of ResNet18. By contrast, we finetune these weights in the RL training.

The extracted features vectors are the input for a gated recurrent unit (GRU) cell, of which the size of the hidden state is 512. With the embedding of the current state, we feed the hidden states of the GRU cell into a fully connected module. With a softmax layer, the output of the RL module indicates the action probability distributions. The agent then samples an action. For example in Fig.~\ref{fig:nn_structure}, it will sample a value for $z_{3,4}$. Recall that there are five identical images for the first step. So we will sample five (possibly different) values for $z_{3,4}$.

Next, one obtains five renewed $\mathbf{y}$'s by setting the newly sampled values for $z_{3,4}$.
So these $\mathbf{y}$'s can start to diverge from each other. Next, the process is repeated, i.e., feature extracting, fully connected mapping over the updated hidden state of GRU cell, sampling from distributions given by the softmax layer. The only difference is to randomly sample the values for $z_{4,5}$, not $z_{3,4}$.

Consider the simplified case in which we only need to set two attributes $z_{3,4}$ and $z_{4,5}$. At this very moment, the final $\mathbf{y}^{(k)}$'s with the same $\mathbf{\epsilon}$ determine the final generated images. These five (probably different) images are input into the aesthetics score network, generating five scores for RL training.

\subsection{Soft policy gradient + self-critical}
\label{sec:soft_gradient}
The standard objective in reinforcement learning is the expected sum of rewards as follows $\sum_t\mathbb{E}_{(\mathbf{s}_t,\mathbf{a}_t)\sim\rho_\pi}[r(\mathbf{s}_t,\mathbf{a}_t)]$. 
Most RL algorithms are working on the conventional notion of deterministic optimality. That is, the optimal policy we are searching for is always deterministic. And this holds at least for the circumstances of full observability \cite{haarnoja2017reinforcement}.
However, intuitively the conventional objective does not consider the importance of stochasticity for exploration.
A sufficient amount of randomness is critical for searching for near-optimal policies, preventing premature convergence to suboptimal ones. One way of keeping randomness is including an entropy regularization term (referred to as entropy bonus) in the loss function, like in \cite{mnih2016asynchronous,schulman2017proximal}. Even so, this is not a systematic way of ensuring the stochasticity of the policy. In that way, the entropy term serves as a regularizer, not as one of the objectives of the RL agent.

In this paper, we instead follows the framework of maximum entropy reinforcement learning (see e.g.~\cite{ziebart2010modeling}), which aims for a policy $\pi^{*}$ as follows:
\begin{equation}
	\pi^*=\underset{\pi}{\arg \max} \sum_t\mathbb{E}_{(\mathbf{s}_t,\mathbf{a}_t)\sim\rho_\pi}[r(\mathbf{s}_t,\mathbf{a}_t)+\alpha \mathcal{H}(\pi(\cdot|\mathbf{s}_t)],
	\label{eq:ent_ojb}
\end{equation}
where $\mathcal{H}(\pi(\cdot|\mathbf{s}_t)$ is the entropy, and $\alpha$ is a hyper-parameter controlling the relative importance of $\mathcal{H}(\pi(\cdot|\mathbf{s}_t)$ in the objective.

The next step is to get the gradient of this objective in \ref{eq:ent_ojb} w.r.t.~the policy parameters. In \cite{haarnoja2017reinforcement}, Haarnoja et al.implicitly gave the following form of the gradient (see Eqs.~(27) and (28) in Appendix B of the arXiv version), and this form is explicitly stated in \cite{shi2019soft}. The result is as follows:
\begin{theorem}[Soft Policy Gradient~\cite{haarnoja2017reinforcement,shi2019soft,sutton2011reinforcement}]
	Suppose that the policy $\pi(\cdot|\mathbf{s})$ is differentiable w.r.t.~its parameters $\mathbf{\theta}$. Then,
	\begin{equation}
		\begin{split}
			&\nabla_{\mathbf{\theta}}J_\mathrm{PG}(\pi_\mathbf{\theta})\\
			=&\mathbb{E}_{(\mathbf{s}_t,\mathbf{a}_t)\sim\rho_{\pi_\mathbf{\theta}}}\left[\left(\hat{Q}_{\pi_\mathbf{\theta}}(\mathbf{s}_t,\mathbf{a}_t)-b(\mathbf{s}_t)\right.\right.\\
			&\left.-\alpha(1+\log\pi_\mathbf{\theta}(\mathbf{a}_t|\mathbf{s}_t))\vphantom{\hat{Q}_{\pi_\mathbf{\theta}}(\mathbf{s}_t,\mathbf{a}_t)}\right)
			\left.\nabla_\mathbf{\theta}\log\pi_\mathbf{\theta}(\mathbf{a}_t|\mathbf{s}_t)\vphantom{\left(\hat{Q}_{\pi_\mathbf{\theta}}(\mathbf{s}_t,\mathbf{a}_t)-b(\mathbf{s}_t)-\alpha(1+\log\pi_\mathbf{\theta}(\mathbf{a}_t|\mathbf{s}_t))\right)}\right]\text{,}
		\end{split}
		\label{eq:soft_gradient}
	\end{equation}
	where $\alpha$ is the temperature hyper-parameter controlling the the extent of exploring, $\hat{Q}_{\pi_\mathbf{\theta}}(\mathbf{s}_t,\mathbf{a}_t)$ is an empirical estimate of the Q-value of the policy, and $b(\mathbf{s}_t)$ is a state-dependent baseline, which can be any function as long as it does not vary with the actions.
	\label{thm:soft_gradient}
\end{theorem}

With the theorem above, we follow the approach proposed in \cite{rennie2017self}, i.e., incorporating the self-critical way of training into the soft policy gradient updates. More concretely, $\hat{Q}_{\pi_\mathbf{\theta}}(\mathbf{s}_t,\mathbf{a}_t)$'s are replaced with Monte Carlo estimations over single trajectories. These estimations correspond to the final five aesthetics scores given by the aesthetics score network. This disposes the need for estimations via separate neural networks for predicting $\hat{Q}_{\pi_\mathbf{\theta}}(\mathbf{s}_t,\mathbf{a}_t)$'s. 
For $b(\mathbf{s}_t)$ in \ref{thm:soft_gradient}, it is set as the average of the five scores. This baseline deviates slightly from \cite{rennie2017self} (which uses the baseline score got by the greedy policy under current $\pi_\mathbf{\theta}$) but experimentally it is more effective as shown in \cite{luo2020better}.

In a word, the main difference from conventional self-critical training is applying the way of self-critical on the soft (not the vanilla) policy gradient. In the experiment section, we show this change is critical to ensure sufficient exploration, and thus this guarantees better performance.

\subsection{Aesthetics score network}
We designed a network dedicated to scoring the aesthetics of human faces. We use the lightweight module ECA (Efficient Channel Attention) \cite{2020ECA-Net}, which acts as a plug-and-play block, and we apply it to the proposed network. We adopt EfficientNet-B4 \cite{2019Efficientnet} as a pre-trained model to extract 1,792 features of the image. Then a convolution operation is performed with a kernel size of 3. We normalize the features before using the ReLU activation function. Next,  it is input to the ECA module to obtain a 4-dimensional vector, and we flatten this vector into a 1-dimensional vector after activation and adaptive average pooling. After this, it is input into the regression network, which returns the aesthetics score.

We trained this aesthetic face scoring model above using the SCUT-FBP5500 dataset \cite{liang2018scut}. The SCUT-FBP5500 dataset contains a total of 5,500 frontal faces and their beauty scores, including 2,000 Asian females, 2,000 Asian males, 750 Caucasian females, and 750 Caucasian males. The batch size is 64 for the training, and the initial learning rate is 1e-4. The training consists of 10 epochs.
For validation, we remap the prediction score of our model to $[1,5]$, which is the beauty score range in SCUT-FBP5500. The RMSE (root-mean-square error) of our model is 0.3058, which is better than the best result of 0.3325 presented in the paper of SCUT-FBP5500 \cite{liang2018scut}. The latter adopts a CNN-based ResNeXt-50 architecture.

\section{Experiments}
All the RL models presented in this section use a batch size of 16. Since the RL agent samples five random trajectories, the literal batch size is 80 for each gradient updating step. By fixing the random seed, we maintain the same validation set across all comparisons. Identical to the experiments conducted in the original paper of EigenGAN, for the Adam optimizer \cite{kingma2014adam,loshchilov2017decoupled}, we use a learning rate of 1e-5 with no $L_2$ regularization over the parameters. For the GRU cell, the size of the hidden state is 512, which is the same value for the output of the ResNet18 module. The ResNet18 module is a pre-trained version on ImageNet \cite{deng2009imagenet}. We set the hyperparameter $\alpha$ in \ref{eq:soft_gradient} as 0.01 for soft policy gradient.

\begin{figure*}
	\centering
	\includegraphics[width=.98\linewidth]{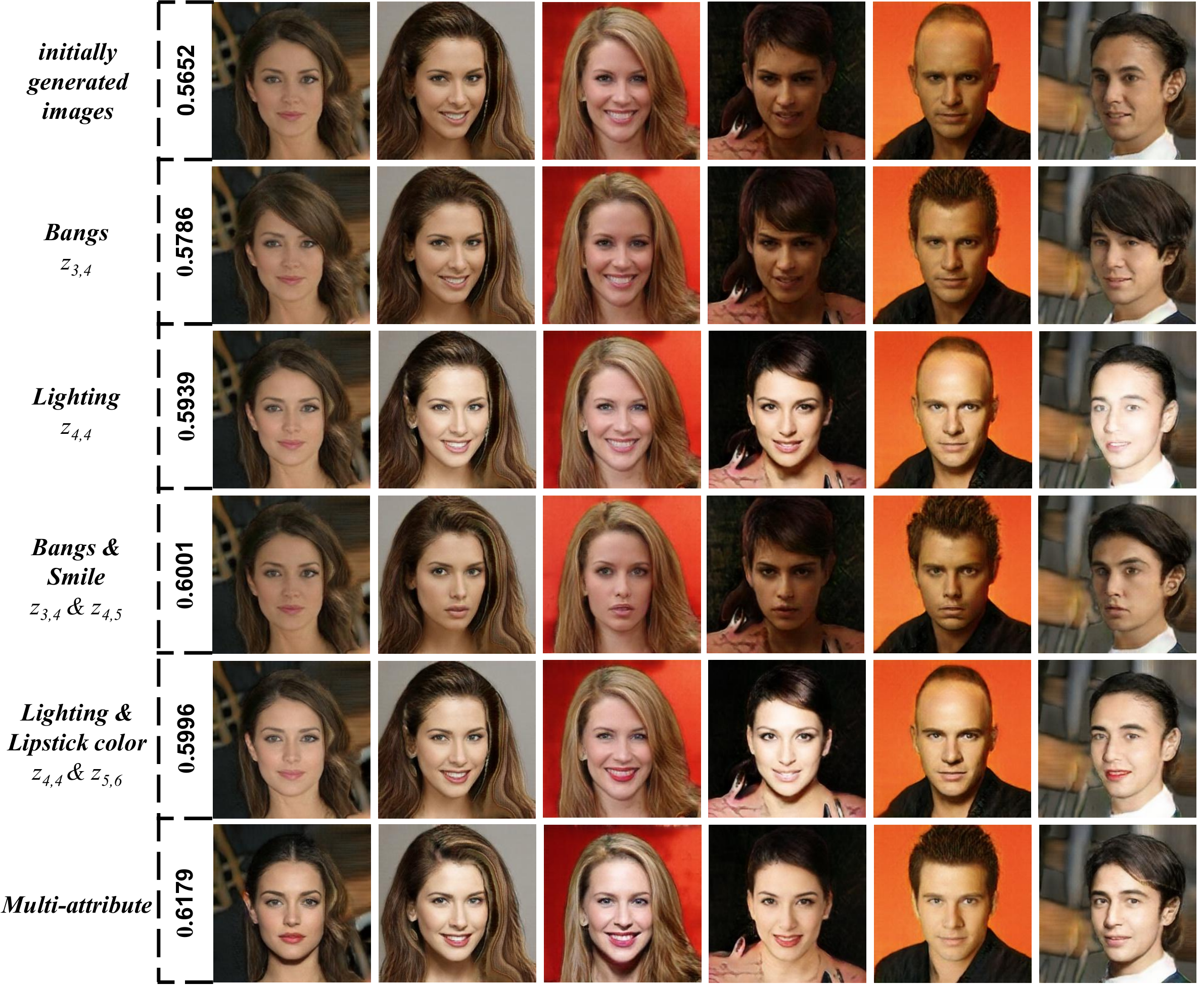}
	\caption{A qualitative comparison of aesthetic improvement with different sets of attributes. Six rows are listed, which indicates six distinct suits of semantics available. Also, next to each row of illustrated pictures, we type vertically the average scores of 1,000 randomly generated test images.}
	\label{fig:exp}
\end{figure*}

For the action space, for each semantic attribute, we discretize the range $[-4.5,4.5]$, taking equally spaced 17 points. It works well, although one could use continuous values or split into more bins. Nonetheless, we observe no significant improvement with more fine-grained splits, partially because the generated images are insensitive to small fluctuations of attribute values. The final models are versions with 20,000 updating iterations, and we report the models with the highest scores on the validation set (please refer to Fig.~\ref{fig:exp}).
\label{sec:exp}

\subsection{Attribute selection}
EigenGAN discovers various facial semantic attributes learned in each layer of subspace, and we wish to analyze their correlations with the aesthetic quality.
We use Spearman's rank correlation coefficient or Spearman's $\rho$, which measures the dependence between two variables by performing linear correlation analysis on the ranks of two variables. For the analysis here, the aesthetics score range is $[0,1]$. The searching range of semantic attributes is $[-4.5,4.5]$. We discretize it with step size 0.5.  We calculate Spearman's $\rho$ for 30 randomly generated pictures. Correlation coefficients greater than 0.8 indicate high correlation; We interpret those with values less than 0.3 as irrelevant. In addition, we perform a significance test over these coefficients. Hypotheses as to this test are $H_0:\rho=0$ and $H_1:\rho\neq 0$. We set the threshold of indicating significant difference as 0.05 for the $p$-values. We list the results in Table~\ref{table:corr}.

Lastly, we have to choose the dimensions with better decoupling effects, i.e., one only observes continuous changes of a \emph{single} semantic attribute when traversing the coefficients of those eigen-dimensions. This step is necessary mainly because, in EigenGAN, different eigen-dimensions might control the same semantic. For example, changes in $z_{5,2}$ and $z_{5,3}$ both affect the hair color. To break the ties, we adopt a greedy policy by choosing the more relevant one. To further narrow the scope, only the attributes with strong correlation ($\vert\rho\vert > 0.3$) and also being significant ($p$-value$<0.01$) are selected. We further filter the candidate set by observing whether they preserve the facial identities or not. After this step, we choose only five semantics, as marked boldly in Table~\ref{table:corr}.

\setlength{\tabcolsep}{4pt}
\begin{table}
	\begin{center}
		\caption{Correlation test results of 30 pictures randomly generated by EigenGAN. The notation $z_{i,j}$ indicates the $j^{th}$ dimension of the $i^{th}$ layer.  When the $p$-value is less than $10^{-4}$, we treat it as 0.0. For repeated attributes, we choose the most relevant one. The final selected ones are marked in bold.}
		\label{table:corr}
		\begin{tabular}{ccrr}
			\hline\noalign{\smallskip}
			D & Attribute & Spearman's $\rho$ & $p$-value\\
			\noalign{\smallskip}
			\hline
			\noalign{\smallskip}
			$z_{2,5}$ & \textbf{Body Pose} & 0.418  & 0.0       \\
			$z_{3,4}$ & \textbf{Bangs} & -0.857 & 0.0           \\
			$z_{3,5}$ & Hair Style & -0.125 & 2.875$\times10^{-3}$   \\
			$z_{3,6}$ & Body Side & 0.188  & 0.0 \\
			$z_{4,1}$ & Pose & 0.328  & 0.0            \\
			$z_{4,2}$ & Background & -0.978 & 0.0      \\
			$z_{4,3}$ & Pose & 0.027  & 0.1551  \\
			$z_{4,4}$ & \textbf{Lighting} & -0.561 & 0.0        \\
			$z_{4,5}$ & Smiling & 0.786  & 0.0         \\
			$z_{4,6}$ & Face Shape & -0.720 & 0.0      \\
			$z_{5,1}$ & Background & -0.198 & 0.0 \\
			$z_{5,2}$ & Hair Color & -0.164 & 0.0 \\
			$z_{5,3}$ & Hair Color & 0.685  & 0.0       \\
			$z_{5,4}$ & Lighting & -0.061 & 0.1484   \\
			$z_{5,5}$ & \textbf{Gaze}& 0.585  & 0.0                 \\
			$z_{5,6}$ & \textbf{Lipstick Color} & 0.471  & 0.0   \\ 
			\hline
		\end{tabular}
	\end{center}
\end{table}
\setlength{\tabcolsep}{1.4pt}

\subsection{Effect of soft policy gradient on entropy}
\begin{figure*}
	\centering
	\subfigure[]{
		\begin{minipage}[b]{0.47\textwidth}
			\centering     
			\includegraphics[width=\linewidth]{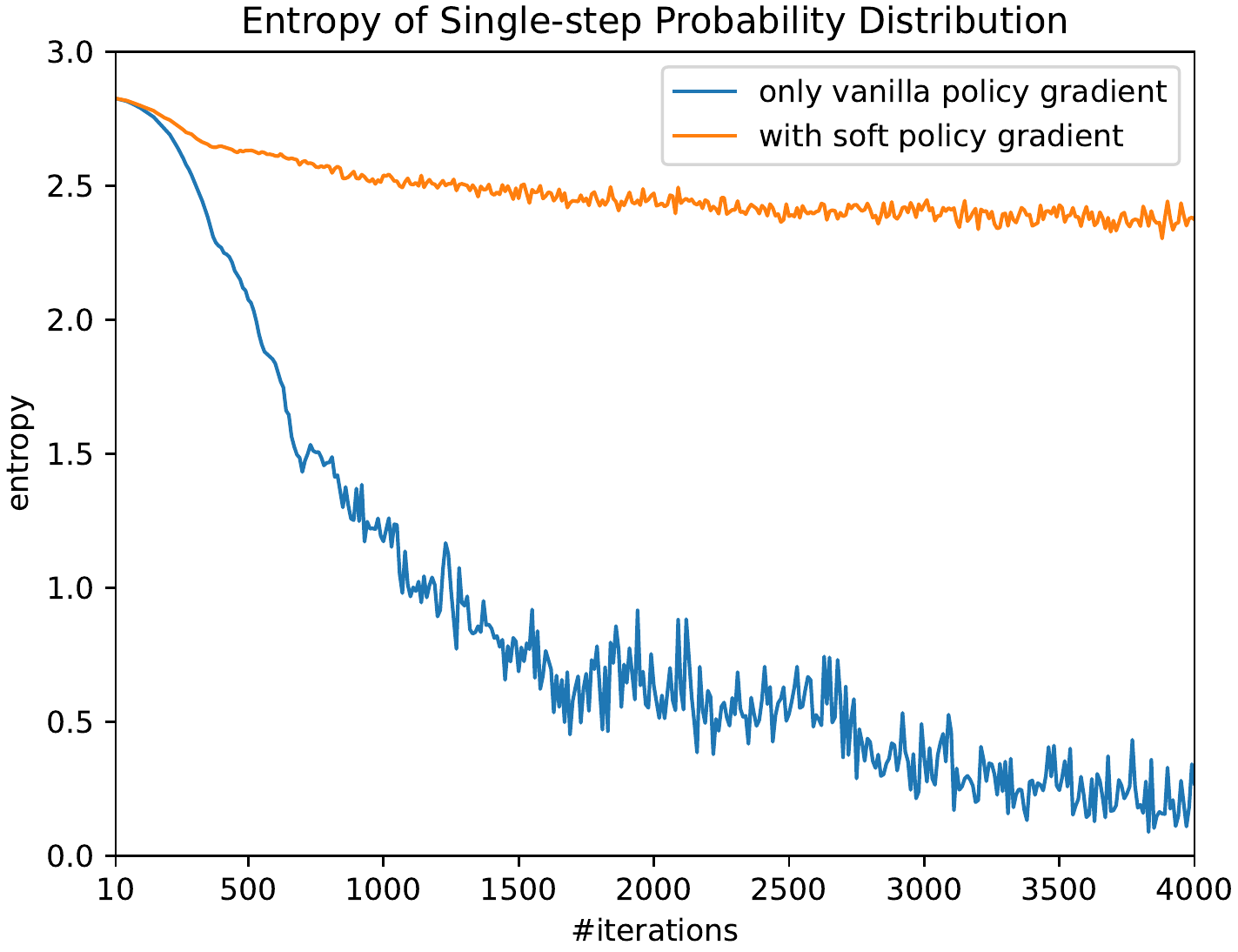}
			\label{fig:entropy}
		\end{minipage}
	}
	\subfigure[]{
		\begin{minipage}[b]{0.47\textwidth}
			\centering     
			\includegraphics[width=\linewidth]{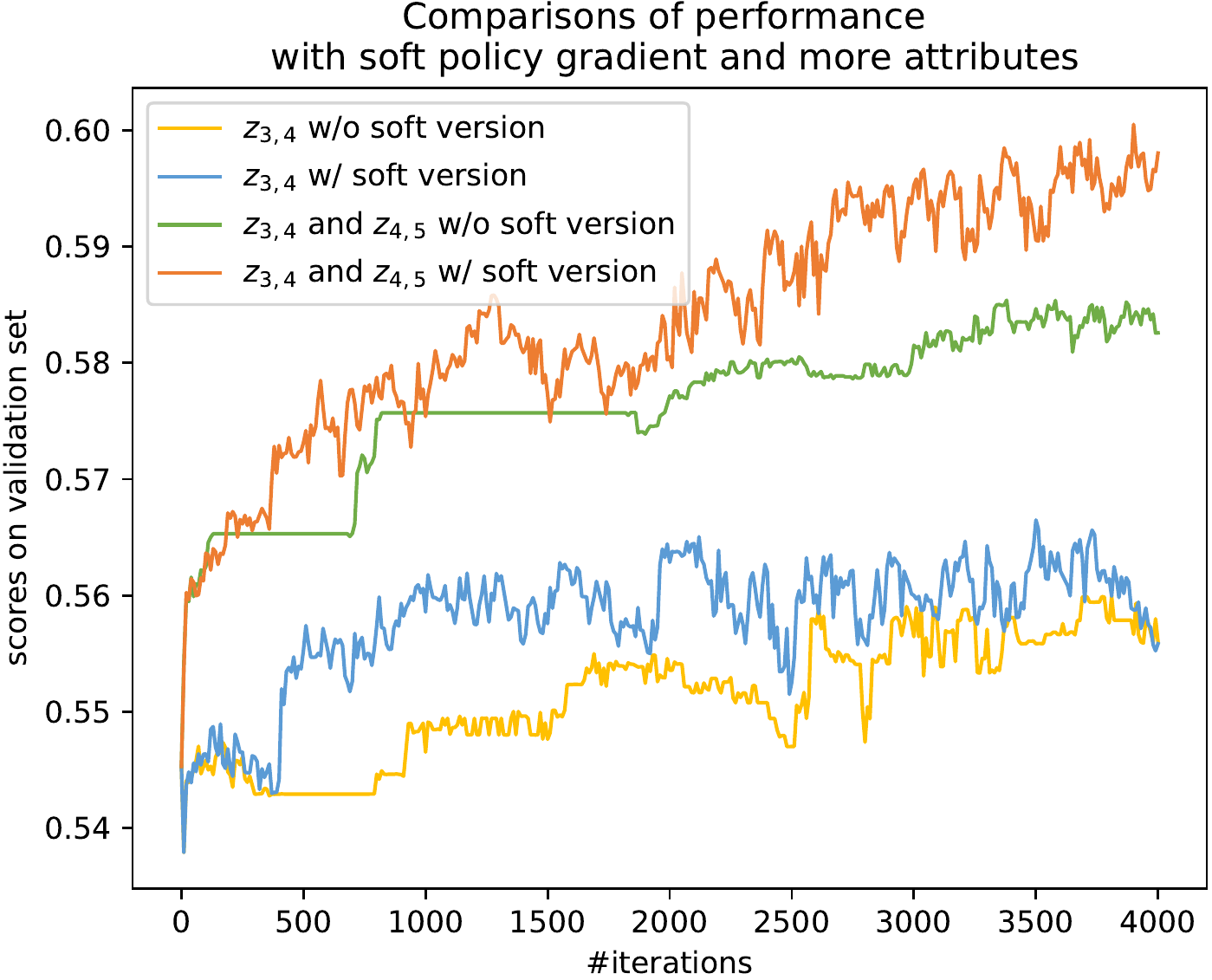}
			\label{fig:bonus}
		\end{minipage}
	}
	\caption{\subref{fig:entropy} Changing of entropy w.r.t.~the number of iterations. \subref{fig:bonus} Performance w.r.t.~the number of iterations when adjusting more attributes or using soft policy gradient.}
\end{figure*}

In \ref{sec:soft_gradient}, we emphasize the role of soft policy gradient in ensuring the higher entropy of $\pi_\mathbf{\theta}$. Without the soft policy gradient, the main phenomenon observed is the quick diminishing of entropy, which indicates the possibility of exploration of diverse choices is gradually becoming zero. For RL agents, this might induce premature convergence to suboptimal deterministic policies.
To simplify the observation, we only consider the case of setting only one dimension, i.e., $z_{3,4}$. Thus it is a single-step environment now. The difficulty of achieving higher scores by changing only $z_{3,4}$ justifies the choice of this dimension. Fig.~\ref{fig:entropy} depicts the changing of entropy w.r.t.~training iterations. The initial entropy value is slightly higher than 2.82. As can be observed, if following vanilla policy gradient updates, the average entropy of distributions defined by $\pi_\mathbf{\theta}(\cdot|\mathbf{s}_t)$ decreases to values less than 0.3, whereas, by using soft policy gradient, this value is kept still above 2.3. Higher entropy values indicate the fact that there are still possibilities for exploration. It helps the agents to learn and perform better finally.

\begin{figure}
	\centering
	\includegraphics[width=0.93\linewidth]{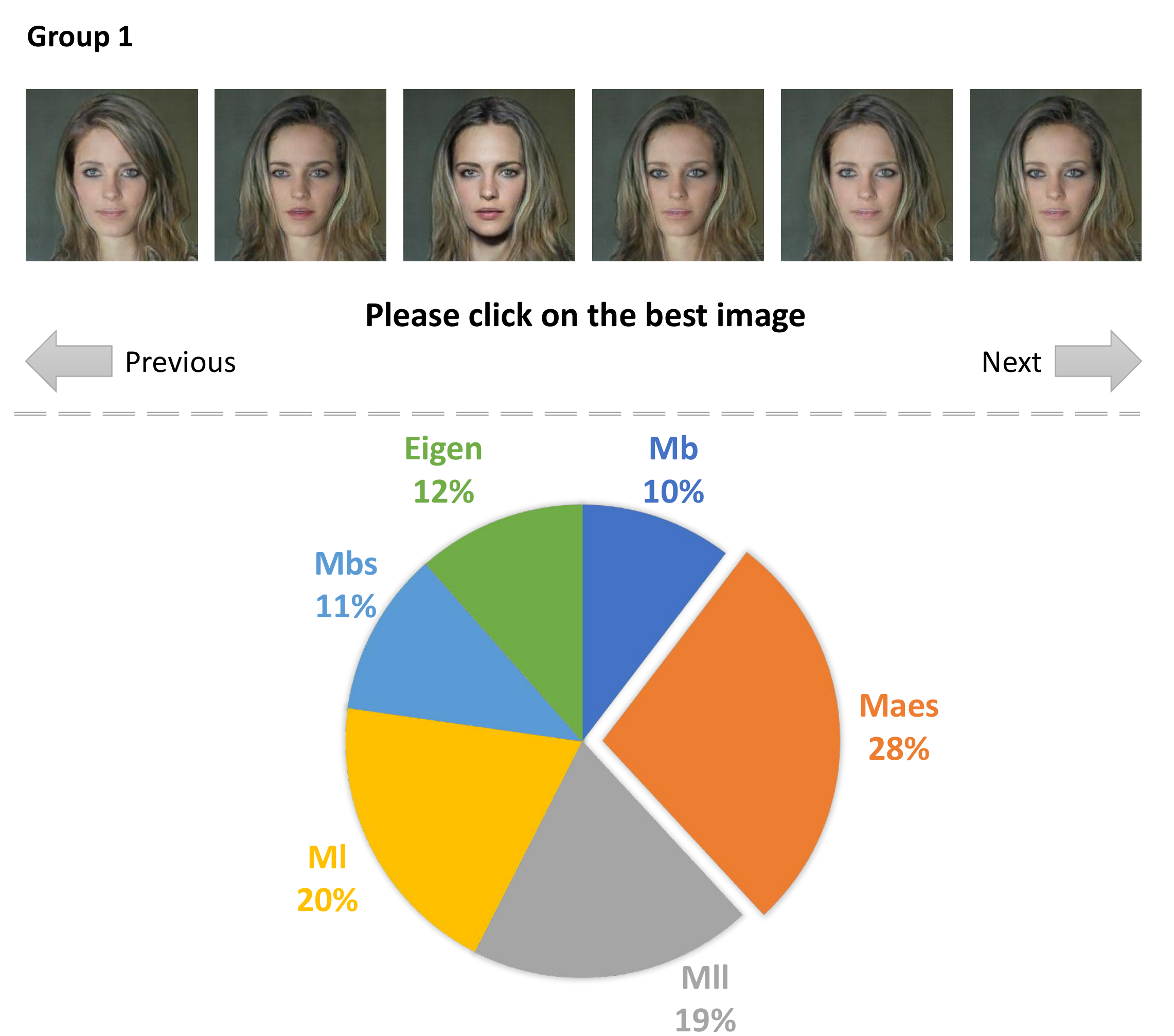}
	\caption{Top: Schematic of subjective experiment. Bottom: Participants' selection statistics for five variants of the proposed model. By fixing the latent variables, these five variants, namely $M_\text{b}$, $M_\text{l}$, $M_\text{bs}$, $M_\text{ll}$, and $M_\text{aes}$, generate 100 groups, 600 images in total. Participants chose the best-looking among each group. $M_\text{aes}$ receives the highest votes.}
	\label{fig:sub}
\end{figure}

\subsection{Improved performance with soft policy gradient and more attributes}

In this section, we extend the former one-step environment to a counterpart of two steps, i.e., $z_{3,4}$ and $z_{4,5}$. So one can observe the improved performance brought by considering more attributes adjustable. Meanwhile, we also show the results on using soft policy gradient or not. Intuitively higher entropy enables the possibility of still exploring the environment and thus getting higher scores. We verify this intuition in the experiment. We note higher entropy of $\pi_\mathbf{\theta}(\mathbf{a}_t|\mathbf{s}_t)$ does not necessarily mean the agent cannot perform optimally. By contrast, as in the inference phase, we only select $\mathbf{a}^{*}={\arg \max }_{\mathbf{a}_t} \pi_\mathbf{\theta}(\cdot|\mathbf{s}_t)$, so as long as preferred actions still dominate the probability mass, intuitively the agent still performs optimally. We adopt a randomly generated set of 80 images for validation. The original average score of these images is 0.54530, shown in Fig.~\ref{fig:bonus} for iteration 0. Consider the increase of scores by adjusting one more attribute. If using vanilla policy gradient,
for one-step setting, the validation score is improved to 0.55991; for two-steps, this value is 0.58539. Meanwhile, for soft policy gradient, these two values are 0.56649 and 0.60051 resp. So this indicates higher scores with more attributes adjustable. Also, most of the time, the curves concerning soft policy gradient are above the curves of vanilla policy gradient, no matter the environment is one-step or two-steps. This observation verifies the advantage of the soft version of policy gradient.

\subsection{Experimental analysis of combinations of attributes}
We evaluate five variants of the proposed method, called $M_\text{b}$ (\emph{bangs}), $M_\text{l}$ (\emph{lighting}), $M_\text{bs}$ (\emph{bangs \& smile}), $M_\text{ll}$ (\emph{lighting \& lipstick color}), and $M_\text{aes}$ (\emph{joint control over full five attributes}).
Fig.~\ref{fig:exp} shows the qualitative comparison of the face images obtained by these five variants, which are tested over
a total of 1,000 images, and six of which are presented as an illustration.
Most images changed by these variants are distinct from the original ones. Moreover, they are all with promoted average aesthetics scores. 
We use FID to evaluate the quality of these images. The FID value of the image generated by $M_\text{aes}$ is 47.11, and the FID value of the initially generated image is 41.07, indicating that the attribute adjustment does not make the quality better, just makes them more beautiful.
Also, Fig.~\ref{fig:exp} shows that controlling the brightness achieves a higher average score than altering the bangs, which indicates that the enhancement differs with distinct semantics. In the fourth row, we observe mouthes of the faces change significantly, from bright smiles to slightly pursed mouthes. Intuitively, as grinning faces have been scarce in SCUT-FBP5500, this might give misleading reward signals to the RL agent. In addition, as Asian faces dominate SCUT-FBP5500 up to 72\%, the aesthetics score network tends to prefer darker hair colors, and hence we remove this semantic in $M_\text{aes}$. Also, we observe $M_\text{aes}$ alters the images in a way, not via a straightforward composing of the changes in each separate single-step model. For example, for the first column, the image generated by $M_\text{aes}$ is different from several other variant models in terms of bangs, lighting, lipstick color, body pose, and gaze. For $M_\text{aes}$, we observe that for lighting, it tends to choose the butterfly light, which highlights the characters and reduces the environmental impact. Some changes are not in line with common perceptions, such as some male faces choosing reddish lipstick colors, partly because the generator lacks gender perception.

\subsection{Subjective analysis}
Because of the subjectivity of image aesthetics, humans show diverse characteristics in their aesthetic thinking.
Even on commonly viewed attractive ones, people might differ subtly in their preferences. We, therefore, conduct a subjective comparative experiment on the five variants presented in the previous section.
We recruit eight users to participate in the subjective study. All users do not know the experiment contents in advance.
In the subjective study, we compared the face images obtained by EigenGAN and our five variants, namely $M_\text{b}$, $M_\text{l}$,$M_\text{bs}$, $M_\text{ll}$ , and $M_\text{aes}$. We show 100 groups of images to all participants. The images generated by the same latent variables are a group, and we display each group of images in random order. The user selects the best image by clicking the screen without limiting the selection time.
Fig.~\ref{fig:sub} shows that $M_\text{aes}$ has obtained the most votes, which indicates its effective controlling of various attributes of the face.

\section{Conclusions}
In this paper, by applying a novel variant of the ingredients in RL, we propose an RL-based model which can learn 
to precisely control the semantics underlying the generator of EigenGAN.
We experimentally show the resulting model can alter the images in an aesthetically more attractive way.
In future work, we will verify the efficacy of the proposed method on real-world image editing. A straightforward way for this task is to first recover the latent codes in EigenGAN corresponding to the input images.

\textbf{Limitations.} The work presented in this paper largely depends on how well EigenGAN can discover disentangled semantics and can reveal these interpretable factors explicitly in the generators. However, we have to manually observe the continuous changes by traversing the coefficient of specific eigen-dimensions. This step helps us filter out these semantics (such as gender, age, race, etc.) that drastically alter the identities of the original images. Another drawback of the method here is the quality of RL policy largely depends on the facial beauty prediction network, this might misleading the RL agent.

\begin{acks}
	This work is partially supported by the National Natural Science Foundation of China (62072014 \& 62106118), the CAAI-Huawei MindSpore Open Fund (CAAIXSJLJJ-2021-022A), the Open Fund Project of the State Key Laboratory of Complex System Management and Control (2022111), the Project of Philosophy and Social Science Research, Ministry of Education of China (20YJC760115), and the Advanced Discipline Construction Project of Beijing Universities (20210051Z0401).
	We gratefully acknowledge the support of MindSpore, CANN and Ascend AI Processor used for this research.
\end{acks}

\bibliographystyle{ACM-Reference-Format}
\bibliography{sample-base}

\end{document}